\documentclass[11pt]{article}

\usepackage{amsmath, amssymb, enumerate, fullpage}
\usepackage{graphicx}
\usepackage{mdframed, xcolor}

\usepackage{algpseudocode}

\usepackage{listings} 	

\usepackage{hyperref}
\hypersetup{
     colorlinks=true,
     linkcolor=blue,
     filecolor=blue,
     citecolor = black,      
     urlcolor=blue!90,
     }

\newcommand{\beginPM}{\begin{pmatrix}}
\newcommand{\closePM}{\end{pmatrix}}

\newcommand{\ds}{\displaystyle}
\newcommand{\realR}{{\mathbb{R}}}
\newcommand{\sphereS}{{\mathbb{S}}}
\newcommand{\calA}{{\mathcal{A}}}		
\newcommand{\calN}{{\mathcal{N}}}
\newcommand{\actiong}{{\mathsf{g}}}		
\newcommand{\groupG}{{\mathsf{G}}}		
\newcommand{\groupE}{{\mathsf{E}}}		

\newcommand{\inner}[2]{\big\langle #1, \, #2 \big\rangle}
\newcommand{\norm}[1]{{\left\Vert #1 \right\Vert_2}}

\newcommand{\myor}{\qquad \mbox{or} \quad}
\newcommand{\mywith}{\qquad \mbox{with} \quad}
\newcommand{\myforany}{\qquad \mbox{for any} \quad}

\newcommand{\bx}{{\boldsymbol x}}
\newcommand{\by}{{\boldsymbol y}}

\newcommand{\bq}{{\boldsymbol q}} 	

\newcommand{\bv}{{\boldsymbol v}}
\newcommand{\bw}{{\boldsymbol w}} 	

\newcommand{\ba}{{\boldsymbol a}}

\begin{document}

\setlength{\baselineskip}{12.8pt}

\title{ Bucketed PCA Neural Networks with Neurons Mirroring Signals	
        \thanks{Contact Email: jhshen@alum.mit.edu; URL: alum.mit.edu/www/jhshen }           }

\author{ Jackie Jianhong  Shen 		\\ \\
		 Financial Services			\\
         New York City, USA 	\\
        \\
        \\
    	\today
        }

\date{}

\maketitle

\begin{abstract}
The bucketed PCA neural network (PCA-NN) with transforms is developed here in an effort to benchmark deep neural networks (DNN's), for problems on supervised classification. Most classical PCA models apply PCA to the entire training data set to establish a reductive representation and then employ non-network tools such as high-order polynomial classifiers. In contrast, the bucketed PCA-NN applies PCA to individual buckets which are constructed in two consecutive phases, as well as retains a genuine architecture of a neural network. This facilitates a fair apple-to-apple comparison to DNN's, esp. to reveal that a major chunk of accuracy achieved by many impressive DNN's could possibly be explained by the bucketed PCA-NN (e.g., 96\% out of 98\% for the MNIST data set as an example). Compared with most DNN's, the three building blocks of the bucketed PCA-NN are easier to comprehend conceptually - PCA, transforms, and bucketing for error correction. Furthermore, unlike the somewhat quasi-random neurons ubiquitously observed in DNN's, the PCA neurons resemble or mirror the input signals and are  more straightforward to decipher as a result.

\vskip 16pt

\noindent \textbf{Keywords:} \; {\small Interpretable AI, Supervised Learning, Neurons, DNN, Mirroring,  PCA, Transforms, Bucketing, Error Correction, MNIST Data, Benchmark Models, SR 11-7. } 
	
\vskip 24pt
\begin{mdframed}[backgroundcolor=blue!10] 
\noindent \textbf{Attention:} \; The current work is designed to be  published exclusively in the Social Sciences Research Network (SSRN) or arXiv preprint server. Its commercial or open-journal publication is prohibited without the prior consent of the author.
\end{mdframed}

\end{abstract}	
	

\newpage
\tableofcontents


\newpage
\section{Introduction}
\label{sec:1-intro}

\newcommand{\wwwSRModelValid}{https://www.federalreserve.gov/supervisionreg/srletters/sr1107.htm}
\newcommand{\wwwMNIST}{https://en.wikipedia.org/wiki/MNIST_database}
\newcommand{\wwwKalmanFilter}{https://en.wikipedia.org/wiki/Kalman_filter}

We propose a neural network system directly based on the principal component analysis (PCA) for supervised classification. The three main components are:
\begin{enumerate}[(a)]
\item direct neuron construction via PCA,

\item neuron transformation, and

\item error correction via bucketed PCA.
\end{enumerate}
They are universally applicable to problems in supervised classification, though the second component on transformation may have to depend on the specific characteristics of the input signal classes.

There are two primary objectives of developing such non-mainstream neural networks.
\begin{enumerate}[(A)]
\item To benchmark mainstream deep neural networks (DNN)~\cite{Goodfellow2016-etal-BK-DeepLearning,LeCun98-Bottou-etal-ATCL}. Traditional parametric models, e.g., linear or logistic regressions, can also offer benchmarking to DNN's but are often less insightful due to major differences in architecture. PCA neural networks (PCA-NN) explored in this work are genuine neural networks although constructed very differently from the mainstream DNN's. Model benchmarking is a crucial tool for effective model validation, e.g., as in the dogmatic supervisory statement ``\href{\wwwSRModelValid}{SS 11-7}" by the United States Federal Reserve Board, for validating all models adopted by the financial industry in the USA. 

Practitioners are often impressed by the high accuracy rates of DNN's but also keep wondering how much can be directly interpreted away from the somewhat black-box nature of DNN's. The current work shows that at least up to 96.00\% of the accuracy rate can be interpreted by the bucketed PCA-NN framework, as applied to the celebrated \href{\wwwMNIST}{MNIST digit data set}. 

\item To construct neural networks with neurons that are much easier to decipher. This is pertinent to the new initiative of creating ``interpretable AI." The author, however, is not in a position to defend or denounce such informal movement. 

In the financial industry for instance, from many Chief Risk Officers for Models (CRO-M's) to front desk quantitative traders, the demand for some level of interpretability is almost ubiquitously on the table. Should a trader sell  a very illiquid municipal bond at the price tag of \$94.15 as projected by a DNN for a notional of \$600 MM? When the entire open market has not observed or reported a single trade for three days, can a DNN  speak a bit of human language to explain how \$94.15 has been derived? When the financial stake is high, the human mind seems to habitually seek comfort from rationality and interpretability. 

The neurons in a typical DNN often appear quasi-random in terms of spatial or correlation patterns, and do not possess the look and feel that resonate with the input signals or their intrinsic characteristics~(e.g., see Figure~\ref{fig:dnn-neurons}). In contrast, the PCA neurons naturally mirror the common structures (or the most significant as expressed by variances or principal values) of the input signals and their population variations.
\end{enumerate}

PCA~\cite{Strang98-BK-IntroLinAlg,Strang19-BK-LinAlgLearningData} has found numerous applications in machine learning. In most classical works (e.g.,~\cite{Ott76-ATCL-Quadratic,schurmann78-ATCL-Word-Recog}), PCA offers a universal way to project complex ensembles of signals onto lower-dimensional feature spaces. Further classification schemes are then applied in these simplified spaces, e.g., quadratic or more general polynomial classifiers. The PCA-NN explored herein differs in three aspects.
\begin{enumerate}[(1)]
\item First, one primary objective of the PCA-NN is to preserve the genuine architecture of a neural network. This allows apple-to-apple benchmarking with DNN's.  As a result, PCA is not merely used as a tool to project onto lower-dimensional feature spaces, but also as a tool to produce interpretable neurons. These neurons can then be transformed under a generic framework. 

\item The second salient feature of the PCA-NN is that PCA is not applied outright to the entire training set, but only to individual ``buckets" and in two sequential phases. The second phase is of particular interest since it achieves the function of error correction, as analogous to error correction in \href{\wwwKalmanFilter}{Kalman filters}.  

\item Finally, the PCA-NN requires no extra non-network classifying schemes such as the polynomial classifiers. PCA-NN maintains the genuine architecture of a neural network, i.e., only involving weighted sums (as a classical model of perceptual neurons), univariate ``activation" functions, and max and argmax. 
\end{enumerate}

In the current work, the bucketed PCA-NN has been illustrated entirely on the MNIST data set of handwritten digits. It achieves a final accuracy score of $96.00^+$\%, but with three fully interpretable components of design. its performance on other data sets may depend on the specific characteristics of the signal classes. The framework of PCA-NN, however, should be universally applicable to classification problems. At a minimum, it shall always provide a genuine benchmarking neural network to DNN's, as well as unveil a major chunk of the accuracy rate achieved by DNN's. 

For the rest of the paper, the three components in the bucketed PCA-NN and their performance will be presented in full details. 

Finally, we emphasize that the current work is only based on low-level vision, i.e., relying only on pixels and their gray values. More robust classification frameworks have to be based on high-level vision toolkit such as hierarchical structuring or context-driven grammars (e.g.,~\cite{Mumford10_Desolneux_BK_PatternTheory}).   

\section{Some Quasi Motivations from Neuroscience}
\label{sec:2-motivations}

One important aspect of an interpretable neural network is to construct neurons that are easy to understand. Such neurons usually capture or mirror certain characteristics of the input signal classes.

Two results in neuroscience \textit{might} support the effort of seeking neurons or network designs that mirror or are directly ``shaped" by stimulus signals. Such a view, however, remains more speculative than scientific unless further neuroscientific evidences are unveiled. 

\newcommand{\wwwMirrorNeurons}{https://en.wikipedia.org/wiki/Mirror_neuron}
\newcommand{\wwwHubelWiesel}{https://en.wikipedia.org/wiki/Simple_cell}
\newcommand{\wwwHeaviside}{https://en.wikipedia.org/wiki/Heaviside_step_function}

\begin{enumerate}[(1)]
\item \href{\wwwMirrorNeurons}{Mirror neurons or systems} are ones that can be activated at the sight or hearing of other people's selected actions, though it is still a developing and debating subject. Actions like yawning, for example, are surprisingly contagious among people confined in a small space. 

Such scenes are often highly selective among hundreds of thousands of image frames projected onto the retinas each day. The almost automated and subconscious mirroring reactions may suggest that the human brain codes and stores these specific scenes or related features directly, instead of relying on generic neural wiring or systems. 

One speculative analogy is to store a unit feature vector $\bw = (w_1, \ldots, w_m) \in \realR^m$ in the canonical Euclidean space, as a neuron or its weights. When any (potentially pre-processed) unit vector $\bv = (v_1, \ldots, v_m) \in \realR^m$ emerges as a stimulus signal, the magnitude of the response $z$ via the Euclidean inner product - 
\[
	z = \inner{\bw}{\bv} = \sum_{j=1}^m w_j * v_j,
\]
can immediately indicate if the pattern $\bw$ has been observed in $\bv$ (say, via the shifted \href{\wwwHeaviside}{Heaviside activation} $H(z - (1-\delta))$ for some tolerance threshold $\delta$). This is done in the genuine fashion of an artificial neuron, i.e., via a weighted sum and an activation modulation. More remarkably, the ``neuron" $\bw$ stores, encodes, or mirrors the type of signals it is supposed to capture.   

\item The discoveries of the Nobel Laureates Hubel and Wiesel~\cite{Hubel-Wiesel59-ATCL-SingleNeuron} on the behavior of the \href{\wwwHubelWiesel}{simple cells} of cats. Unlike the quasi-random patterns of most artificial/computational neurons, these neurons are only activated by certain \textit{regular} spatial patterns, i.e., by specific orientations and spatial oscillations of input stimuli. The summation behavior (on activation and prohibition) in their visual fields suggests that these neurons more or less mirror the edges, edgelets, or linear oscillations that exist ubiquitously in natural scenes and are crucial bottom-up visual cues for high-level visual recognition~\cite{ChanShen2005-BK-Image,Nitzberg-Mumford93-BK}. Such findings support the speculation that neurons can be  ``shaped" by nature and may directly encode or mirror frequent and important external stimuli.
\end{enumerate}

\newcommand{\wwwDataNIST}{https://en.wikipedia.org/wiki/MNIST_database}

\section{Data Set}
\label{sec:3-data}

The framework of the bucketed PCA-NN is universally applicable to problems on supervised classification. To develop the main ideas, we focus on the celebrated \href{\wwwDataNIST}{digit-recognition data set by NIST} (\textit{National Institute of Standards and Technology}, USA). Due to some modifications, it is now commonly referred to as the MNIST data set. 

The MNIST data set is summarized as follows.
\begin{enumerate}[(1)]
\item Each data point is a monochrome image of 28 by 28 pixels, capturing a handwritten digit from $\{ 0, 1, \ldots, 9 \}$.

\item There are totally 60,000 train images, and 10,000 test images.

\item In Python, based on Tensorflow and Keras it can be loaded via a command line like:
\begin{lstlisting}
    from keras.datasets import mnist

    (trainImages, trainLabels), (testImages, testLabels) \
    = mnist.load_data()
\end{lstlisting}

\end{enumerate}
Using two layers of neurons of sizes (400, 10) with full connectivity and some other common network features, a DNN could achieve an impressive accuracy rate of 98.15\% for the test data set.

\begin{figure}[!htbp]
\centering
\includegraphics{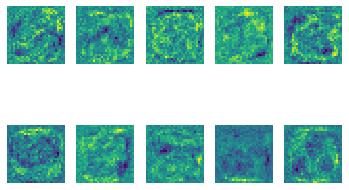}
\caption{ The first 10 neurons of a typical DNN implementation for MNIST. The somewhat quasi-random patterns resulted from weights optimization of DNN do not relate well to the input signal classes, e.g., the 10 digits in the current study.}
\label{fig:dnn-neurons}
\end{figure}

The first 10 neurons for such a DNN implementation are plotted as monochrome images in Figure~\ref{fig:dnn-neurons}, using weights as pixel values. As common to DNN's, the individual neurons do not relate intuitively well to the input handwritten digits. They demonstrate some quasi-random patterns produced by the weights optimization procedure. 

\section{The Bucketed PCA Neural Network}
\label{sec:4-PCANN}

As mentioned earlier in the Introduction, readers should be aware of the difference from the earlier PCA works (e.g.,~\cite{Ott76-ATCL-Quadratic,schurmann78-ATCL-Word-Recog}), where PCA components are first constructed from the \textit{entire} training data set and then a high-order polynomial classifier is deployed based on the reductive PCA representation. In the current work, PCA is applied to each bucket of training samples, and bucketing takes two consecutive steps. Furthermore, it also intends to maintain the authentic architecture of neural networks in order to make fair apple-to-apple comparison with DNN's.  

To summarize the entire effort, the current work helps reveal how much of DNN's impressive accuracy rates can be interpreted by classical frameworks like the present, e.g., at least 96.0\% out of 98.0\% for the MNIST data set. 

\subsection{Raw PCA Neurons and Neural Networks}
\label{subsec:4.1-PCA-neurons}

\subsubsection{PCA of Digit Images}
\label{subsub:4.1.1-PCA}
Let $\Omega$ denote the collection of training samples, e.g., the 60,000 digit images in MNIST. A generic individual sample is denoted by $\bx$. Let $\alpha \in \calA $ denote all supervised labels or classes, i.e., 
\[
	\bx \in \Omega \longrightarrow \alpha = \phi(\bx) \in \calA,
\]
where $\phi$ denotes the supervision. For MNIST, $\calA$ consists of the 10 single digits:
\[
	\calA = \{\text{`0'}, \text{`1'}, \ldots, \text{`9'} \}.
\]

Given a supervision $\phi$, the sample set can be partitioned into single-label buckets $\Omega_\alpha$'s:
\[
	\Omega = \bigcup_{\alpha \in \calA} \Omega_\alpha, \mywith
	\Omega_\alpha = \{ \bx \in \Omega \mid \phi(\bx) = \alpha \}.
\]

For MNIST, each sample data point is a monochrome 2D image of a handwritten digit:
\[
 \bx \in \Omega \subset [0,1]^{28 \times 28} \subset \realR^{28 \times 28}.
\]
A pixel value $x_{ij} = 1$ stands for the brightest gray scale while $x_{ij} =0$ for the dimmest. The background against which digits are handwritten is typically zero or close to. 

Since the perception of a digit is gray-scale invariant, in the current work each image $\bx \in \Omega$ is first normalized to a unit vector:
\[
	\bx \longrightarrow \frac{\bx}{\norm{\bx}}, \mywith
	\norm{\bx}^2 = \mathrm{trace}( \bx \bx^T ) = \sum_{i,j} x_{ij}^2. 
\]
In this way, all sample digit images in MNIST live on the unit hyper-sphere $\sphereS^{28\times 28 -1}$. This helps improve both data visualization and the interpretibility of the PCA approach (See Figure~2 for a visual illustration).

\begin{figure}[!htbp]
\centering
\includegraphics[scale=0.60]{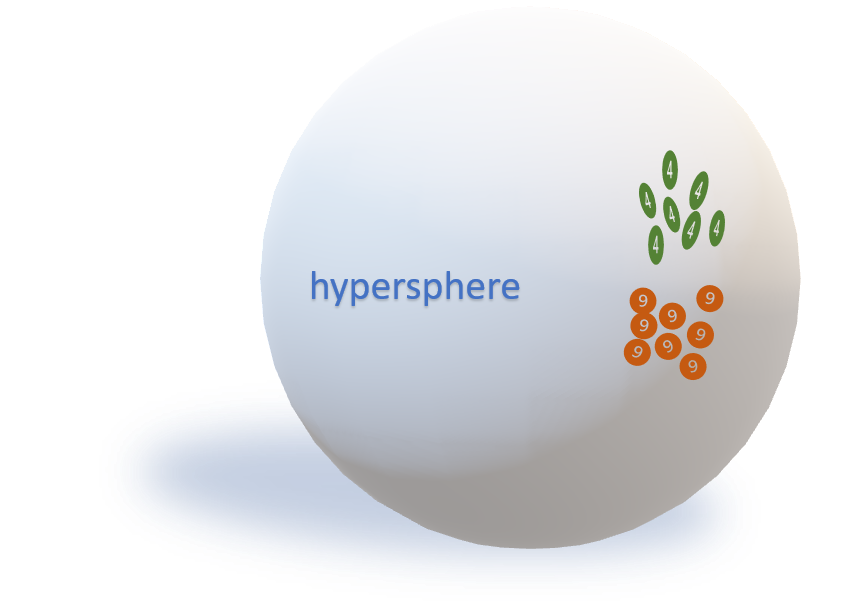}
\caption{ After normalization, all digit images live on the unit hyper-sphere $\sphereS^{28\times 28 - 1}$. }
\label{fig:spherical-view}
\end{figure}

\newcommand{\wwwLinAlgStrang}{http://math.mit.edu/~gs/linearalgebra/}

In order to call the numerical \href{\wwwLinAlgStrang}{SVD decomposor} in Python:
\begin{lstlisting}
	numpy.linalg.svd(),
\end{lstlisting} 
one linearly stretches the 2D image arrays to 1D row vectors via:
\[
	\bx \longrightarrow X = \bx.\mathrm{reshape}(28 \times 28 ).
\]	
For any given digit $\alpha \in \calA$, the associated sample subspace $\Omega_\alpha$ can be numerically enumerated as a $|\Omega_\alpha| \times (28\times 28)$ matrix:
\[
	M_\alpha = \beginPM
	X_1 \\
	X_2 \\
	\vdots \\
	X_{|\Omega_\alpha|}
	\closePM
\]
Calling the SVD decomposer, the subspace is decomposed via:
\[
	M_\alpha = U^T \cdot D \cdot V.
\]
Here $D = \mathrm{Diag}(\sigma_1, \ldots, \sigma_{28\times 28})$ stores all the singular values in descending order, and $V$ stores an orthonormal basis  in $\realR^{28 \times 28}$ (as row vectors for convenience):
\[
	V  = \{ V_i \mid 1 \le i \le 28 \times 28 \}.
\]
Notice that $|\Omega_\alpha| \gg 28\times 28$ for MNIST. Finally, for any image sample $\bx$ or $X$ in $\Omega_\alpha$,
\[
	X  \in \mathrm{span}\{ V_1, \ldots, V_{28 \times 28} \}.
\]

\subsubsection{PCA Neurons and Networks}

As typical for many real-world signals, the singular values decay to zero fast, which gives rise to the most common way of weeding out noises from real signal structures~\cite{Strang98-BK-IntroLinAlg,Strang19-BK-LinAlgLearningData}. Let $r \in (0, 100\%)$ denote a \textit{tail} truncation parameter, e.g., $r = 20\%$. For a given PCA with $K$ descending singular values, define the $r$-truncation index: $K_r = K$ if $\sigma_K^2 > r$, and otherwise,
\[
	K_r = \min \left\{ k: \; \sum_{i=k+1}^K \sigma_i^2 \le r \right\}.
\]

For digit classification, let $K_r(\alpha)$ denote the $K_r$ index for the PCA associated with the matrix $M_\alpha$ as defined above (for $\Omega_\alpha$). Then define the neuron set under a given truncation level $r$ by:
\begin{equation}
\label{eqn:neuron-set-N-alpha}
	\calN_\alpha = \left\{ \bv_k = V_k.\mathrm{reshape}((28,28)) :\; 1 \le k \le K_r(\alpha) \right\}.
\end{equation}
Each neuron in $\calN_\alpha$ is generically denoted by $\bv^{(\alpha)}_k$ to indicate the $\alpha$-dependency. Also the neurons have been reshaped to the original image dimension of $28$ by $28$ pixels. 

The raw PCA neural network (PCA-NN) is built upon all these neurons:
\[
	\calN = \bigcup_{\alpha \in \calA} \cal N_\alpha.
\]
The total number of neurons is:
\[
	|\calN| = \sum_{\alpha \in \calA} |N_\alpha| = \sum_{\alpha \in \calA} K_r(\alpha),
\]
under a given tail truncation parameter $r$. 

The PCA-NN is comprised of the following two layers.
\begin{enumerate}[(A)]
\item (\textbf{First Layer})  On the frontal layer directly facing an input image sample $\bx$, the response of neuron $\bv^{(\alpha)}_k$ is the weighted sum or the canonical Euclidean inner product:
\begin{equation}
\label{eqn:response-inner-product}
	z_k^{(\alpha)} = \inner{\bx}{\bv^{(\alpha)}_k} = \sum_{1 \le i, j\le 28} v^{(\alpha)}_{k,ij} \cdot x_{ij}.
\end{equation}
For each such neuron, the activation function is a square: $\psi(z)  = z^2$. 

\item (\textbf{Second Layer})
The second layer consists of $|\calA|=10$ aggregating neurons: for each $\alpha \in \calA$, the neuron $u_\alpha$ computes the following response:
\begin{equation}\label{eqn:u-alpha-ssq}
	u_\alpha = \sum_{1\le k \le K_r(\alpha)} 1.0 * \psi\left( z_k^{(\alpha)} \right)
				= \sum_{1\le k \le K_r(\alpha)} \inner{\bx}{\bv^{(\alpha)}_k}^2.
\end{equation}
In term of weights on the outputs of the first layer, it can also be written as:
\begin{equation}\label{eqn:pca-second-layer}
	u_\alpha = \sum_{\beta \in \calA, 1\le k\le K_r(\beta) } w_{\beta, k}^{(\alpha)} \cdot \psi\left( z_k^{(\beta)} \right),
\end{equation}
where $\ds w_{\beta, k}^{(\alpha)} =1.0$ when $\beta = \alpha$, and 0 otherwise.

The final classifier for a given sample $\bx$ is directly defined by the argmax:
\begin{equation} \label{eqn:argmax-classifier}
	\hat{\alpha}(\bx) = \mathrm{argmax} \{ u_\alpha :\; \alpha \in \calA \}.
\end{equation}

\end{enumerate}

\begin{figure}[!htbp]
\centering
\includegraphics[scale=0.72]{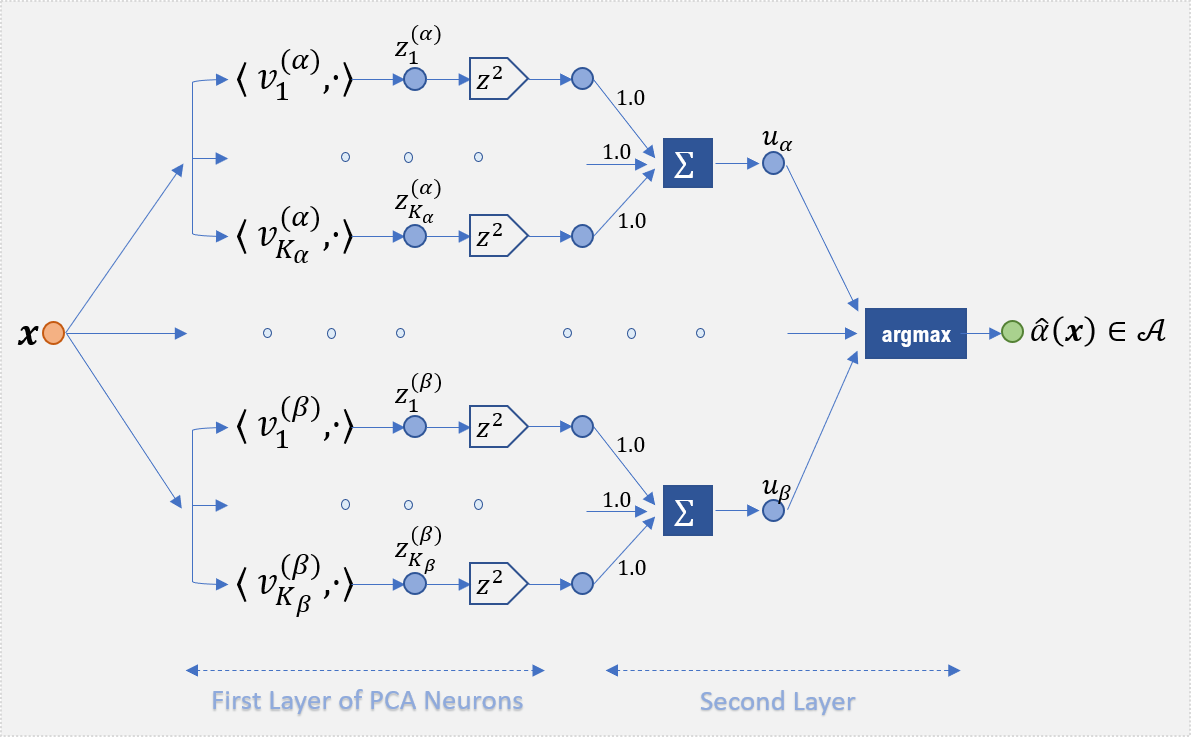}
\caption{ The raw PCA Neural Network (PCA-NN). When the PCA tail truncation is set to  $r=20.0\%$ for instance, the average $K_r(\alpha)$ is only about 13 (neurons) for MNIST. In contrast with most classical PCA works, PCA-NN applies PCA to each bucket, instead of the entire training set. It also relies only on the genuine neural network aggregator or synthesizer $\max$ or $\mathrm{argmax}$, instead of separate high-order polynomial classifiers. Also stay tuned that these raw buckets will be further refined.}
\label{fig:pcann-raw}
\end{figure}

\subsubsection{Performance of the Raw PCA Neural Network}
\label{subsub:4-1-3-performance}

We call such a neural network the \textit{raw} PCA-NN. The performance of the raw PCA-NN is moderate.
\begin{enumerate}[(a)]
\item The (out-of-sample) accuracy rate for the 10,000 MNIST test images is 93.49\%.
\item The (in-sample) accuracy rate for the 60,000 MNIST training images is 93.50\%.
\end{enumerate}
While the performance may seem less appetizing compared with DNN's, the raw PCA-NN does possess the genuine 2-layer neural network structure and hence offers an apple-to-apple benchmark to the DNN's and their impressive performance. It shows at least that a major chunk (e.g., 93.50\% out of 98.0\%+) of DNN's performance can be easily interpretable. This will be further improved later on. 

The raw PCA-NN does not explicitly rely on forward or backward optimization. Instead, it only relies on the PCA decomposer, which itself is certainly an implicit classical optimizer - seeking the least-dimensional subspaces where the samples are maximally concentrated. For instance, in the case of setting a tail truncation level at $r = 20\%$ (i.e., retaining the 80\% of the principal spectra), the average dimension for the 10 digits is only about 13, i.e., seeking 13 neurons and their spanned spaces to efficiently represent the key features of about 6,000 samples for each digit. 

\begin{figure}[!htbp]
\centering
\includegraphics[scale=0.96]{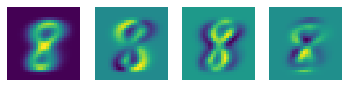}
\caption{ The first 4 PCA neurons $\bv_{1:4}^{(\alpha)}$ for digit $\alpha=8$. Unlike the quasi-random patterns of neurons in most DNN's (e.g., Figure~\ref{fig:dnn-neurons}), these neurons are easily interpretable. The first one (from the left) bears the highest principal value and captures the average or most common features of the given digit. It corresponds to the clustering center of a given digit population on the hyper-sphere, as illustrated in Figure~\ref{fig:spherical-view}.  The second one captures the \textit{angular} or \textit{rotational} variations in handwriting. The third neuron captures \textit{horizontal} displacement or thickening in handwriting, while the fourth neuron captures \textit{vertical} displacement or variation.}
\label{fig:neurons-examples}
\end{figure}

The first 4 PCA neurons for digit ``8" have been plotted in Figure~\ref{fig:neurons-examples} and can be easily interpreted as demonstrated in the caption.

\subsection{Neuron Transforms}

\subsubsection{Signal Transforms}
\label{subsub:4-2-1-signal-transforms}

Neuron transform is a way to recycle existing neurons for extracting further signal features. Convolutional neural network~\cite{Goodfellow2016-etal-BK-DeepLearning}, for instance, is a well-known example in which a base neuron or set is transformed via a series of translations to participate in the same network.

Transforms are certainly signal specific, and different signal classes may embrace different transforms. Following the earlier notation, let $\Omega$ denote the signal sample space, $\bx \in \Omega$ a generic sample, and $\calA$ the label set. Let $\phi: \Omega \to \calA$ denote the ground-truth supervision or classification. A classification is said to be \textit{invariant} under a transform group (or collection) $\groupG$, if  
\begin{equation}\label{eqn:invariance}
	\phi(\actiong \bx) = \phi(\bx), \myforany \bx \in \Omega, \; \actiong \in \groupG.
\end{equation}

\newcommand{\wwwEuclideanGroup}{https://en.wikipedia.org/wiki/Euclidean_group}
For digit recognition, the 2D \href{\wwwEuclideanGroup}{Euclidean group} $\groupE$ over $\realR^2$ naturally comes to mind. Let $\bq =(z_1, z_2)$ denote an arbitrary pixel in $\realR^2$. Any pair of data
\[		
	\actiong=(\ba, Q) = (\ba \in \realR^2,\quad Q\; \mbox{a 2D orthogonal matrix} )
\]
naturally defines a transform on any given image $\bx = \bx(\bq)$ by:
\[
	\actiong \bx (\bq) := \bx( Q^{-1}\bq + \ba), \qquad \mbox{i.e., moving the image in rigid motions.}
\]
But common sense reminds us the following.
\begin{enumerate}[(i)]
\item Digit recognition is not reflection invariant. These three digits $\{0, 1, 8\}$ are approximately reflection invariant under horizontal or vertical reflections. But all the rest are not. 
\item Digit recognition is rotationally invariant only to \textit{certain degrees}. For instance, when the digit ``6" is rotated by 180 degrees, it actually becomes another digit ``9." Similarly, the digit ``1" only becomes the Mandarin character for ``1" after a 90-degree rotation.
\end{enumerate}
In addition, the MNIST data set has largely been normalized by framing all digits around the center of the image domain. Hence translation (via $\ba$ in the above definition) also would not make much contribution. Such reckoning suggests that one should focus only on small rotations $Q=R_\theta$, with $|\theta| \le U$:
\[
	R_\theta \bq = R_\theta(z_1, z_2) = (z_1 \cos \theta - z_2 \sin \theta, \; z_1 \sin \theta  + z_2 \cos \theta).
\]
This translates to the rotation operator $R_\theta$ on any image sample $\bx = \bx(z_1, z_2)$:
\begin{equation}\label{eqn:rotation}
	R_\theta \bx(\bq) := \bx(R_{-\theta}\bq),
	\myor
	R_\theta \bx (x, y) : = \bx(z_1 \cos \theta + z_2 \sin \theta, \; -z_1 \sin \theta  + z_2 \cos \theta). 
\end{equation}
For instance, $R_{20^\circ}\bx$ rotates an image $\bx$ by $20^\circ$ counterclockwise. 

The angle range $U$ can be empirically defined, e.g., $U = 30^\circ$. In reality, people's writing habits, including left-right handedness and hand-paper alignment, result in the natural dispersion of the orientations of a given digit within a \textit{moderate} range. 

\subsubsection{Duality and Neuron Transforms}
\label{subsub:4-2-2-neuron-transforms}

For the purpose of discussion, it is assumed that image samples $\bx, \by$ are defined on the entire 2D visual field $(z_1, z_2) \in \realR^2$. The reception of an input image $\bx$ by a neuron $\bv$ is defined by the weighted sum, or more formally, the inner product in $L^2(\realR^2)$:
\[
	\inner{\bx}{\bv}	= \int_{\realR^2} \bx(z_1, z_2) \bv(z_1, z_2) \; dz_1 dz_2 
						= \int_{\realR^2} \bx(\bq) \bv(\bq) \; d\bq.
\]
The duality for bounded linear operators refers to:
\begin{equation}\label{eqn:linear-duality}
	\inner{R_\theta \bx}{\bv} 
		= \int_{\realR^2}\bx\left( R_{-\theta} \bq \right) \bv(\bq) d\bq
		= \int_{\realR^2}\bx\left( \hat \bq \right) \bv( R_\theta \hat \bq) d\hat\bq
		=\inner{\bx}{R_{-\theta} \bv},
\end{equation}
or simply $R_\theta^\ast = R_{-\theta}$ in $L^2(\realR^2)$. Such duality is 
standard in functional analysis but can be very useful for neural network design for the following reasons.
\begin{enumerate}[(a)]
\item There is no need in the network to perform direct transforms on input signals. Instead, under the duality principal, it is sufficient to only insert new neurons transformed from existing ones.

\item The duality principal even extends to more general transforms:
\[
	\actiong \bx(\bq) : = \bx( \Phi^{-1} \bq ),
\] 
where $\Phi(\cdot)$ is a diffeomorphism of the 2D visual field $\realR^2$ (e.g., in the context of deformable  templates in~\cite{Mumford10_Desolneux_BK_PatternTheory}). It can be easily shown that:
\begin{equation} \label{eqn:diffeomorphism}
	\inner{\actiong \bx (\bq)}{\bv(\bq)} 
	=  \inner{ \bx(\hat \bq) }{ \actiong^\ast \bv( \hat \bq ) }, \mywith
	\actiong^\ast \bv(\hat \bq) :=  \bv( \Phi \hat \bq) \cdot J_{\hat \bq} \Phi,
\end{equation}
where $J$ is the Jacobian volume multiplier in $\realR^2$. Then to properly capture  $\Phi$-transformed image signals, it is sufficient to only insert the dual neurons $\actiong^\ast \bv$ into the existing neural networks. 
\end{enumerate}

\subsubsection{Transformed PCA-NN}
\label{subsub:4-2-3-transformed-PCA}

In actual numerical implementation, once a set of transforms $\actiong \in G$ have been chosen, following the earlier settings in Eqn.~\eqref{eqn:neuron-set-N-alpha},~\eqref{eqn:response-inner-product}, and~\eqref{eqn:pca-second-layer}, one can augment the raw PCA-NN by the following approach.
\begin{enumerate}[(a)]
\item For each transform $\actiong$, under duality each neuron $\bv_k^{(\alpha)}$ is transformed to a new one $\ds \actiong^\ast \bv_k^{(\alpha)}$,  denoted by $\ds \bv_k^{(\actiong, \alpha)} $ for convenience. Then each neural set $\calN_\alpha$ defined in Eqn.~\eqref{eqn:neuron-set-N-alpha} is transformed accordingly to $\calN^{(\actiong)}_\alpha$. 

\item For a given sample signal $\bx \in \Omega$, following Eqn.~\eqref{eqn:response-inner-product} and~\eqref{eqn:pca-second-layer}, one obtains the responses of the first and second layers:
 \[
 	z_k^{(\actiong, \alpha)}, \quad u_\alpha^{(\actiong)}. 
 \]
 
\item Then for each label or class $\alpha \in \calA$, the final $\alpha$-score for a given signal $\bx$ is:
\[
		U_\alpha = \max \{ u_\alpha^{(\actiong)} :\; \actiong \in G \}.
\]
For convenience, it is assumed that the transform set $G$ also contains the identity transform so that the original raw PCA-NN response $u_\alpha$ is also included. Then the final classifier $\hat \alpha_G$ is defined similarly as in Eqn.~\eqref{eqn:argmax-classifier}:
\begin{equation}\label{eqn:alpha-classifer-G}
	\hat \alpha_{\tiny G}(\bx) = \mathrm{argmax} \{ U_\alpha :\; \alpha \in \calA \}.
\end{equation}
We call this augmented neural network the $G$-transformed PCA-NN. 
\end{enumerate}

\begin{figure}[!htbp]
\centering
\includegraphics[scale=0.70]{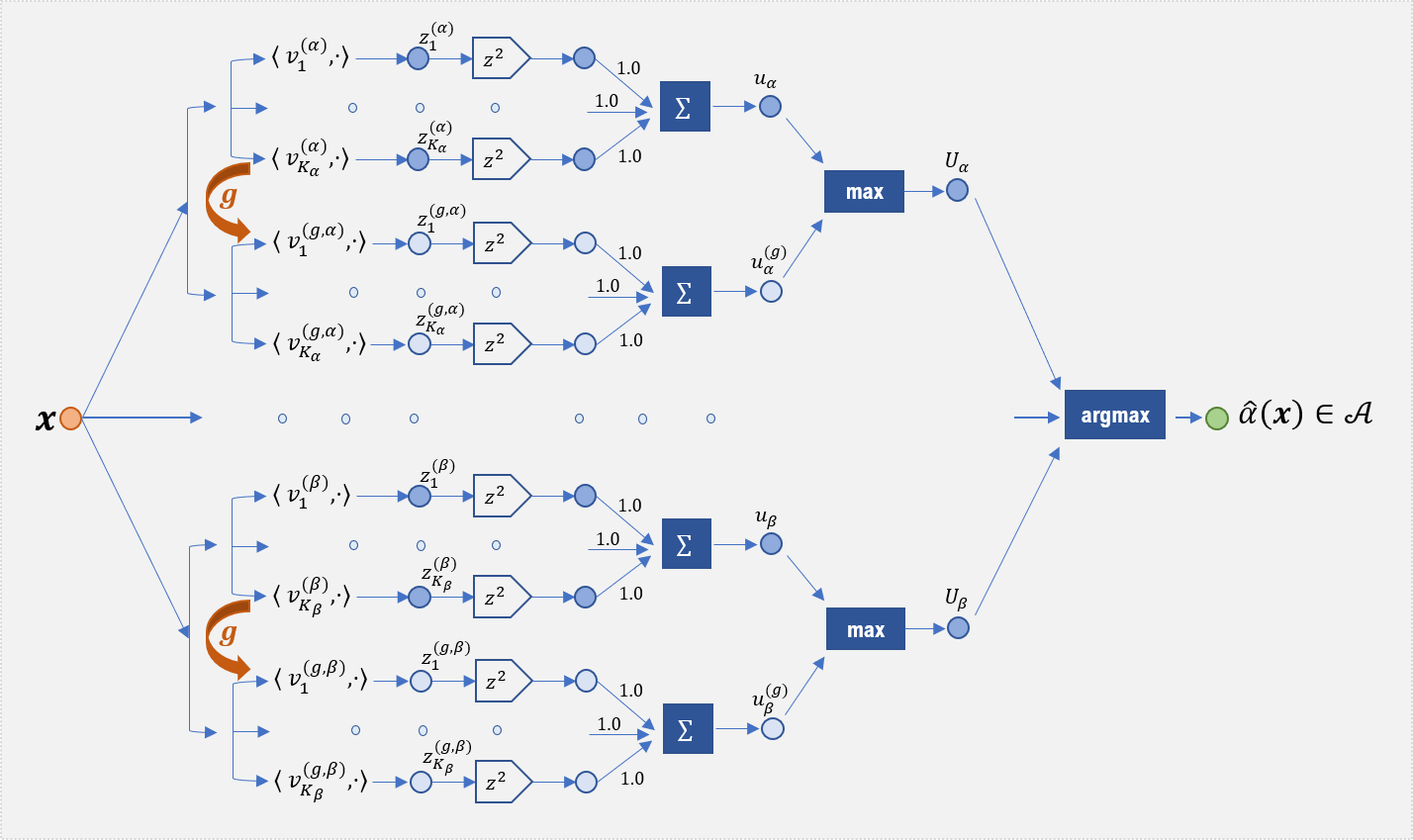}
\caption{ The transformed PCA-NN, as illustrated via a single (non-identity) transform $\actiong \in G$. (See Eqn.\eqref{eqn:alpha-classifer-G} and also compare with Figure~\ref{fig:pcann-raw}.) The transformed neurons can capture transformed input signals without directly turning to inverse transforms on the signals, while maintaining the genuine architecture of a neural network.  }
\label{fig:pcann-transformed}
\end{figure}

\subsubsection{Performance of the Transformed PCA-NN}
\label{subsub:4-2-4-transform-performance-G}

Once a range of angles $(-U, U)$ is set (e.g., $U=30^\circ$ as mentioned earlier), one can discretize it via a chosen gap, e.g., $\Delta \theta = 3^\circ$ or $5^\circ$:
\[
 	\theta_{-M} < \cdots < \theta_{-1} < \theta_0 = 0.0 < \theta_{1} < \cdots < \theta_M.
\] 
Here $\theta_0=0.0$ corresponds to the identity operator, i.e., original raw PCA-NN. It is possible to select the best combination simply via examining the associated classification accuracy. For the present work, we eventually settle to the following transform set:
\[
	G = \left\{ \theta_{-1} = -12.0^\circ,\quad \theta_0 = 0.0,\quad \theta_1 = 12.0^\circ \right\}.
\]
The moderate degree of 12 may indicate the average angular variation among handwritten digits.

The performance improvement is as follows:
\begin{enumerate}[(i)]
\item For training samples (of 60,000), the accuracy is 94.11\%, which is a 61 basis point (bps) improvement from 93.50\% of the raw PCA-NN.

\item For test samples (of 10,000), the accuracy rate is 94.09\%, which is an improvement of 60 bps from 93.49\% of the raw PCA-NN. It is consistent with the in-sample training data.
\end{enumerate} 

Later on we shall see that combined with the last component of bucketing and error correction, the transformation method will eventually contribute to an ultimate accuracy rate above 96.0\% - an improvement of totally 250 basis points from the raw PCA-NN.


\subsection{Bucketing and Error Correction}
\label{subsec:4-3-bucketing}

The third and also the last stage of the PCA-NN design involves bucketing and error correction. 

\subsubsection{Systematic Factors behind Erroneous Buckets}
\label{subsub:4-3-1-erroneous-bucketing}


Error correction is a key step in~\href{\wwwKalmanFilter}{Kalman Filtering}. The underlying philosophy however could be universally applicable, i.e., embedded within prediction errors is usually some useful information about a hidden target and hence such errors can be used for prediction improvement \textit{a posteriori}. The third and also the last component of the PCA-NN follows exactly this spirit.

Let $\alpha, \beta \in \calA$ denote two labels (or digits in the current application). Given a learning network $\calN$ which outputs $\hat\phi(\bx) \in \calA$ for any input sample $\bx \in \Omega$. Let $\phi(\bx)$ denote the supervised labeling or ground-truth classification. When $\hat\phi(\bx) \neq \phi(\bx)$, the network $\calN$ produces a classification error.

Let the bucket $\Omega_{\beta\mid\alpha}$ denote all the training samples that are supervised as $\alpha$ but classified by $\calN$ as $\beta$, i.e.,
\begin{equation} \label{eqn:bucket-alpha-to-beta}
\begin{split}
	\Omega_{\beta \mid \alpha} 
		&= \{ \bx \in \Omega :\; \phi(\bx) = \alpha,\;  \hat\phi(\bx) = \beta \} \\
		&= \{ \bx \in \Omega_\alpha :\; \hat\phi(\bx) = \beta \}. 
\end{split}
\end{equation}
Then for any $\alpha \in \calA$,  $\Omega_{\alpha \mid \alpha}$ is a correct bucket while any $\Omega_{\beta \mid \alpha}$ is an error bucket when $\beta \neq \alpha$. Clearly,
\[	
		\Omega_\alpha = \bigcup_{\beta \in \calA} \Omega_{\beta \mid \alpha}. 
\]
An effective network $\calN$ should see high concentrations on the correct buckets $\Omega_{\alpha \mid \alpha}$'s. 

When the total training population $\Omega$ is substantive (e.g., 60,000 for MNIST), some erroneous buckets $\Omega_{\beta \mid \alpha}$'s could still contain statistically meaningful pools of samples. For digit recognition in MNIST, for instance, $\Omega_{4 \mid 7}$ - all images of digit ``7" that are misclassified as ``4" by the raw PCA-NN, could still contain a few hundreds of samples out of the entire training set. 

\begin{figure}[!htbp]
\centering
\includegraphics[scale=0.80]{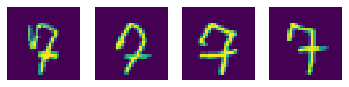}
\caption{ Some examples in the error bucket $\Omega_{4 \mid 7 }$, i.e., those supervised as ``7" but classified as ``4" by the raw PCA-NN. It demonstrates a systematic deviation due to the habit of striking through the digit ``7," a perfect recipe for creating confusion with the digit ``4" (in low-level vision). }
\label{fig:bucket7to4}
\end{figure}

There often exists a systematic driver that leads to so many misclassified samples. Take $\Omega_{4 \mid 7}$ for example. Some people (esp. in Continental Europe) habitually strike through ``7" to differentiate it from the digit ``1" in handwriting. As demonstrated in Figure~\ref{fig:bucket7to4}, in combination with another habit of ``hooking" the top horizontal line of ``7," this could create the perfect recipe for confusion with the digit ``4."

\subsubsection{The Bucketed PCA-NN for Error Correction}
\label{subsub:4-3-2-bucketed-pca}

A general classification problem usually does not attempt to unearth the exact physical causes behind such sizable error buckets. Instead, one holds the general assumption that any sizable error bucket may result from a yet-to-be characterized systematic pattern (e.g., a ``7" with a through-strike), or more generally, a new statistical mode. It implies that the associated label $\alpha \in \calA$ may be \textit{multimodal} and  the raw PCA-NN only captures the main mode. 

This inspires the following bucketed PCA-NN, as improved from the raw PCA-NN. First, the neuron sets for ``sizable" or substantial buckets are constructed.
\begin{mdframed}[backgroundcolor=orange!05] 
\begin{algorithmic}
	\ForAll {label $\alpha \in \calA$}
		\ForAll {predicted label $\beta \in \calA$ (by the raw PCA-NN)}
			\If {$\# \Omega_{\beta \mid \alpha} < M$}
				\State {skip}
			\Else
				\State {construct PCA neuron set $\calN_{\beta\mid\alpha}$ for $\Omega_{\beta \mid \alpha}$}
			\EndIf
		\EndFor
	\EndFor
\end{algorithmic}
\end{mdframed}
The cutoff $M$ is to ignore buckets that are insignificant in terms of size. For the present work, we have chosen $M=20$ heuristically. In general it can be chosen in proportion to the size of the overall training pool. In addition, the bucketed PCA neuron set $\calN_{\beta \mid \alpha}$ is constructed exactly as the raw PCA-NN, with a spectral tail cutoff $r$ (e.g., $r=20\%$ as consistent with the raw). 

For convenience, a generic PCA neuron from a PCA bucket $\calN_{\beta\mid\alpha}$ is denoted by $\bv_k^{(\beta\mid\alpha)}$. 

Also notice that for each label $\alpha \in \calA$, the ``diagonal" bucket $\Omega_{\alpha \mid \alpha}$ contains most of the samples from $\Omega_\alpha$, and its neural set $\calN_{\alpha \mid \alpha}$ captures the prevailing features of the main mode of $\alpha$. 

Once the bucketed neuron sets are constructed, the bucketed PCA-NN classifier is then defined exactly as the raw PCA-NN. 
 
\begin{mdframed}[backgroundcolor=orange!05] 
\begin{algorithmic}
	\State{input a sample $\bx$ to be classified}
	\ForAll {label $\alpha \in \calA$}
		\ForAll {non-empty bucketed PCA neural set $\calN_{\beta \mid \alpha}$}
			\State{calculate the neuron-aggregated response $u_{\beta\mid\alpha}$  (similar to $u_\alpha$ in Eqn.~\eqref{eqn:u-alpha-ssq})}
		\EndFor
		\State{calculate the bucket-aggregated response via: $u_\alpha = \max\{ u_{\beta \mid \alpha} :\; \beta \}$}
	\EndFor
	\State{project the label for $\bx$: \( \hat\alpha(\bx) := \mathrm{argmax}\{u_\alpha :\; \alpha \in \calA \} \) }
\end{algorithmic}
\end{mdframed}

Notice that the bucketed PCA-NN maintains the authentic structure of a neural network, and is structurally identical to the transformed PCA-NN as depicted in Figure~\ref{fig:pcann-transformed}, as long as the neurons $\bv_k^{(\actiong, \alpha)}$\;'s are replaced by the neurons $\bv_k^{(\beta \mid\ \alpha)}$\;'s for each PCA neuron bucket $\calN_{\beta\mid\alpha}$. Hence we have omitted the illustrative figure.

\subsubsection{Performance of the Bucketed PCA-NN, and the Transformed}
\label{subsub:4-3-3-performance-bucketed-pca}

The performance for the bucketed PCA-NN is as follows.
\begin{enumerate}[(i)]
\item For training images, compared with the accuracy of 93.50\% of the raw PCA-NN, the bucketed PCA-NN achieves 96.23\%, with an improvement of 273 basis points.  

\item For test images, compared with the accuracy of 93.49\% of the raw PCA-NN, the bucketed PCA-NN achieves 94.93\%, with an improvement of 144 basis points.
\end{enumerate}

The improvement is more salient for in-sample training images than for the test. After all, the bucketed neurons have been derived from the training images. 

Finally, we apply the transform method of the  preceding subsection to the bucketed PCA-NN. Assuming that the transform set (or group) is $\groupG$, the neural network can then be summarized easily using the same pseudo code for the bucketed PCA-NN. As in the previous discussion, $\groupG$ is assumed to contain the identity transform. 

\begin{mdframed}[backgroundcolor=orange!05] 
\begin{algorithmic}
	\State{input a sample $\bx$ to be classified}
	\ForAll {label $\alpha \in \calA$}
		\ForAll {non-empty bucketed PCA neural set $\calN_{\beta \mid \alpha}$}
			\ForAll {transform $\actiong \in \groupG$}
				\State{calculate the neuron-aggregated response $u^{(\actiong)}_{\beta\mid\alpha}$ for $\actiong$-transformed neuron set $\calN_{\beta\mid\alpha}^{(\actiong)}$}
			\EndFor
			\State{calculate the transform-aggregated response via: 
					$u_{\beta\mid\alpha}=\max\{ u^{(\actiong)}_{\beta\mid\alpha}:\; \actiong \in \groupG  \}$ }
		\EndFor
		\State{calculate the bucket-aggregated response via: $u_\alpha = \max\{ u_{\beta \mid \alpha} :\; \beta \}$}
	\EndFor
	\State{project the label for $\bx$: \( \hat\alpha(\bx) := \mathrm{argmax}\{u_\alpha :\; \alpha \in \calA \} \) }
\end{algorithmic}
\end{mdframed}

We call this final neural network
\[
	\mbox{\textit{the Bucketed PCA-NN with Transforms}.}
\]

Notice that it carries the genuine structure of a neural network, and that the network flow is identical to what is depicted in Figure~\ref{fig:pcann-transformed} - by substituting any raw neuron set $\calN_{\alpha}$ with
its (non-empty) refined buckets $\calN_{\beta\mid\alpha}$\;'s, and also electing $u_\alpha = \max_{\beta}\;\{ u_{\beta\mid\alpha} \}$ as the delegate of label/class $\alpha$ for the final $\mathrm{argmax}$-layer. 

For the bucketed PCA-NN with transforms, the ultimate performance is as follows.
\begin{mdframed}[backgroundcolor=green!05] 
\begin{enumerate}[(i)]
\item For training images, compared with the accuracy of 93.50\% of the raw PCA-NN, the bucketed PCA-NN with transforms achieves 96.99\%, with a total improvement of 349 basis points. 

The transforms (via two rotations) have thus contributed to an improvement of 76 basis points, as compared with 96.23\% of the bucketed PCA-NN (without transforms). 

\item For test images, compared with the accuracy of 93.49\% of the raw PCA-NN, the bucketed PCA-NN with transforms achieves 96.12\%, with a total improvement of 263 basis points.

The transforms (via two rotations) have thus contributed to an improvement of 119 basis points, as compared with 94.93\% of the bucketed PCA-NN (without transforms). 
\end{enumerate}
\end{mdframed}

\section{Conclusion}

To conclude, the current work has demonstrated that a major chunk (i.e., about 96.00$^+$\%) of the impressive accuracy rates  from various neural networks is \textit{interpretable}.  The bucketed PCA-NN has relied on the three interpretable components for network design - PCA, neuron transformation, and error correction via bucketing.  

It is worth emphasizing that our primary objective is to construct genuine neural networks without using non-network frameworks such as high-order polynomial classification (as in other classical PCA-related efforts). This allows a genuine apple-to-apple comparison with neural-network classifiers, including DNN's. 

\section*{Acknowledgments}

The project has been stretched on and off from 2019 to 2021 through a prolonged and challenging period of pandemic and unemployment. The author is very grateful to all the friends and former colleagues who have unconditionally encouraged and supported him at the personal level, esp. Rafa Santander (my former manager and friend at the Santander Bank, USA), Andreza Barbosa (Goldman Sachs),  Andrew Ang (BlackRock), Huazhang Luo, Xiangwei Liu, Dan Pirjol, as well as my dear recruiter Lou Rodriguez. Years later this brief note could become a fragrant rose petal reminding me about the beauty of the seemingly unpredictable dynamics of life and our era.

\end{document}